\documentclass[10pt,twocolumn,letterpaper]{article}

\usepackage{cvpr}              

\usepackage[table]{xcolor}
\usepackage{arydshln}
\usepackage{multirow}
\usepackage{amsfonts}
\usepackage{booktabs}
\usepackage{pifont}
\usepackage{amsmath}
\usepackage{graphicx}

\usepackage{algorithm}
\usepackage{algpseudocode}

\definecolor{cvprblue}{rgb}{0.21,0.49,0.74}
\usepackage[pagebackref,breaklinks,colorlinks,allcolors=cvprblue]{hyperref}

\def\paperID{16504} 
\def\confName{CVPR}
\def\confYear{2026}

\title{SAT: Selective Aggregation Transformer for Image Super-Resolution}

\author{Dinh Phu Tran \\
{\tt\small phutx2000@kaist.ac.kr} \and
Thao Do \\
{\tt\small thaodo@kaist.ac.kr} \and
Saad Wazir \\
{\tt\small saad.wazir@kaist.ac.kr} \and
Seongah Kim \\
{\tt\small kimsa0322@kaist.ac.kr} \and
Seon Kwon Kim \\
{\tt\small lukaskim@kaist.ac.kr} \and
Daeyoung Kim \\
{\tt\small kimd@kaist.ac.kr} \\
School of Computing, KAIST, Republic of Korea \\
}

\begin{document}
\maketitle
\begin{abstract}

Transformer-based approaches have revolutionized image super-resolution by modeling long-range dependencies. However, the quadratic computational complexity of vanilla self-attention mechanisms poses significant challenges, often leading to compromises between efficiency and global context exploitation. Recent window-based attention methods mitigate this by localizing computations, but they often yield restricted receptive fields. To mitigate these limitations, we propose \textbf{S}elective \textbf{A}ggregation \textbf{T}ransformer (SAT). This novel transformer efficiently captures long-range dependencies, leading to an enlarged model receptive field by selectively aggregating key-value matrices (reducing the number of tokens by \textbf{97\%}) via our Density-driven Token Aggregation algorithm while maintaining the full resolution of the query matrix. This design significantly reduces computational costs, resulting in lower complexity and enabling scalable global interactions without compromising reconstruction fidelity. SAT identifies and represents each cluster with a single aggregation token, utilizing density and isolation metrics to ensure that critical high-frequency details are preserved. Experimental results demonstrate that SAT outperforms the state-of-the-art method PFT by \textbf{up to 0.22dB}, while the total number of FLOPs can be reduced by \textbf{up to 27\%}.
Code: \url{https://github.com/PhuTran1005/SAT}.

\end{abstract}    
\vspace{-4mm}

\section{Introduction}

Image super-resolution (SR) is a longstanding challenge in computer vision, aiming to recover high-resolution (HR) images from low-resolution (LR) inputs. As an ill-posed inverse problem, it requires modeling complex LR-HR mappings, where capturing global context is crucial for recovering fine textures and edges. Convolutional
neural networks (CNNs) \cite{dong2015image, kim2016accurate, lim2017enhanced, kim2016deeply, zhang2018residual} have mitigated this challenge by utilizing local kernels to focus on salient features. Yet, their locality limits the ability to exploit global context, resulting in artifacts like blurring or aliasing.
Recently, ViT \cite{dosovitskiy2020vit} has transformed computer vision by enabling global modeling via self-attention, inspiring new directions in SR field.

\begin{figure}[t]
    \centering
    \includegraphics[width=0.49\textwidth, height=4cm]{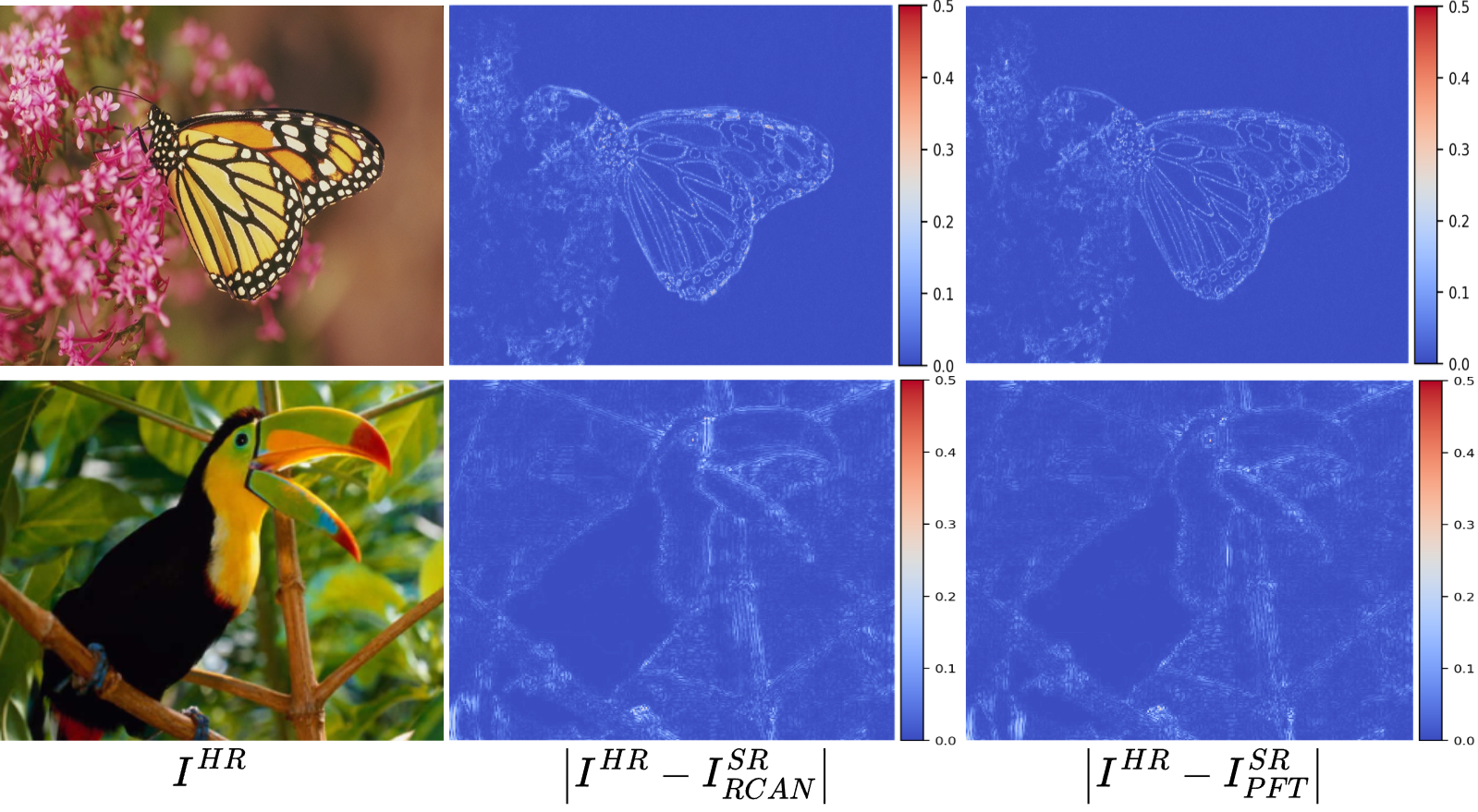}
    \vspace{-7mm}
    \caption{The pixel-wise absolute error between HR and SR images from RCAN \cite{zhang2018image} and PFT \cite{long2025progressive}. These concentrated error areas at high-frequency regions motivate our SAT’s design: we preserve full-resolution Query while compressing Key–Value tokens in homogeneous areas for achieving efficient global attention.}
    \label{fig:motivation}
    \vspace{-6mm}
\end{figure}

Early adopters, such as IPT \cite{chen2021pre}, show the potential of pre-trained Transformers for image SR tasks. Subsequent works \cite{liang2021swinir, tran2024channel, chen2023activating, chen2022cross, zhang2022accurate, tran2025vsrm} use window-based attention and channel attention for enhanced pixel reconstruction. These methods clearly surpass prior CNN-based methods. However, unlike global attention, the local framework restricts attention to a small fixed area. Recently, some works have tried to deal with efficiency and global context exploitation. For instance,
graph-based methods, like IPG \cite{tian2024image},  use flexible local-global graphs to enhance reconstruction.
Still, IPG requires substantial FLOPs and its hardware-unfriendly graph aggregation leads to increased memory usage. ATD \cite{zhang2024transcending} uses an external token dictionary to enhance the attention regions, but leads to an extra FLOPs while introducing limited extra information. PFT \cite{long2025progressive} links attention maps across layers for focused attention. Yet, their error propagation in early layers might degrade overall performance. Moreover, SR inherently requires more computation in high-frequency regions than in smooth areas (see Fig. \ref{fig:motivation}). However, most existing methods employ uniform processing for the entire image, resulting in inefficient allocation of computation. 
Although recent works \cite{chen2021pre, long2025progressive} try to allocate computation efficiently, the imbalance between spatial complexity and computation remains underexplored. 

To bridge these gaps, including restricted receptive field, error propagation, and inefficient resource allocation, we propose Selective Aggregation Attention (SAA). SAA enables efficient global attention by selectively aggregating key-value matrices while preserving query's full resolution. In SAA, Density-driven Token Aggregation (DTA) algorithm identifies and aggregates low-frequency regions in the key-value matrices,
focusing resources on detail-rich areas, thereby reducing significant computation. We then propose Feature Norm Restoration as a post-processing step in DTA to maintain the distribution of feature norms after aggregation process. Consistent feature distribution is crucial for encoding perceptual information \cite{ho2020contrastive} and layer normalization processing \cite{ba2016layer}. SAA primarily focuses on global modeling and can be complemented by a dedicated module for modeling local details. Hence, we integrate SAA into a hybrid Transformer architecture, alternating with local window attention to achieve a complementary global-local structure, further improving the model's performance.

\noindent In summary, this paper makes the following contributions:

\begin{itemize}
\item We propose Selective Aggregation Attention (SAA) as an efficient global attention. SAA is able to capture global dependencies while reducing substantial computations.

\item In SAA, we propose Density-driven Token Aggregation (DTA) for selectively aggregating key-value matrices to reduce the number of tokens by 97\%, while keeping full-resolution query.
DTA efficiently adapts density-peak principles to avoid quadratic complexity in the center selection process, while similarity-weighted aggregation with Feature Norm Restoration preserves semantic coherence and consistent feature norms during aggregation.

\item We provide a comprehensive theoretical analysis, including low-complexity guarantees (Theorem 3.1) and approximation bounds (Theorem 3.2), demonstrating that our method achieves substantial speedup with provable bounds on quality degradation.
 
\item In general, we propose Selective Aggregation Transformer (SAT), which achieves a new state-of-the-art performance in SR, validated through extensive comparisons with all recent methods and rigorous ablation studies.
\end{itemize}
\section{Related Work}

\textbf{Image Super-Resolution.} Deep learning has reshaped the SR field \cite{timofte2018ntire, chen2024ntire, zhang2023ntire}. Some early CNN-based methods, such as SRCNN \cite{dong2015image}, pioneered end-to-end training, and EDSR \cite{lim2017enhanced} designing residual blocks for depth. Attention-enhanced models, such as RCAN's \cite{zhang2018image} channel attention or HAN's \cite{niu2020single} hierarchical attention, improved focus on salient features. Transformers have since dominated: IPT \cite{chen2021pre} utilizes pre-training for restoration tasks, SwinIR \cite{liang2021swinir} uses shifted windows for efficiency, and CAT \cite{chen2022cross}, CPAT \cite{tran2024channel} enhance cross-window interactions and frequency learning. HAT \cite{chen2023activating} uses self- and channel-attention to activate more pixels for better SR quality. However, these methods restricted attention to a limited area. 
Graph-based method, IPG \cite{tian2024image}, uses variable-degree aggregation by treating pixels as nodes in the image graph. Yet, creating this graph remains costly, and hardware-unfriendly graph aggregation increases VRAM usage. ATD \cite{zhang2024transcending} enlarges attention area by using external dictionary tokens and category-based attention. However, this added tokens are limited to approximate global attention while adding more overhead. 
PFT \cite{long2025progressive} links all attention maps across layers to focus on crucial regions. However, early layers may emphasize irrelevant tokens, causing error propagation that can degrade model's performance. PFT also progressively discards other tokens, which still contribute to the SR output. In contrast, 
SAA  efficient global modeling while still utilizing all pixels in the reconstruction process. 



\noindent \textbf{Efficient Attention Mechanisms.} Efficient attention mechanisms \cite{wang2021pyramid, tu2022maxvit, yang2022scalablevit, chen2021regionvit, ali2021xcit} aim to reduce the quadratic complexity of vanilla self-attention \cite{vaswani2017attention, tran2022trans2unet, do2025reference}. PVT \cite{wang2021pyramid} and RGT \cite{chen2023recursive} design a spatial-reduction module using convolution layers to compress feature maps before computing attention. However, PVT remains a high computational cost to balance with performance, while RGT compresses features into a very compact representation, leading to a loss of fine-grained details and struggling with diverse degradations in SR.
MaxViT \cite{tu2022maxvit} proposes the grid attention to gain sparse global attention. ScalableViT \cite{yang2022scalablevit} scales attention matrices from both spatial and channel dimensions. These approaches reduce overall complexity but still lose many fine-grained details that are crucial for SR.
Moreover, XCiT \cite{ali2021xcit} proposes a “transposed” self-attention that operates across channel dimension to reduce complexity. However, it cannot explicitly model the spatial relationship.
Consequently, there is a growing need for an efficient exploration approach to balance performance and computational cost.

\begin{figure*}[t]
    \centering
    \includegraphics[width=0.95\textwidth]{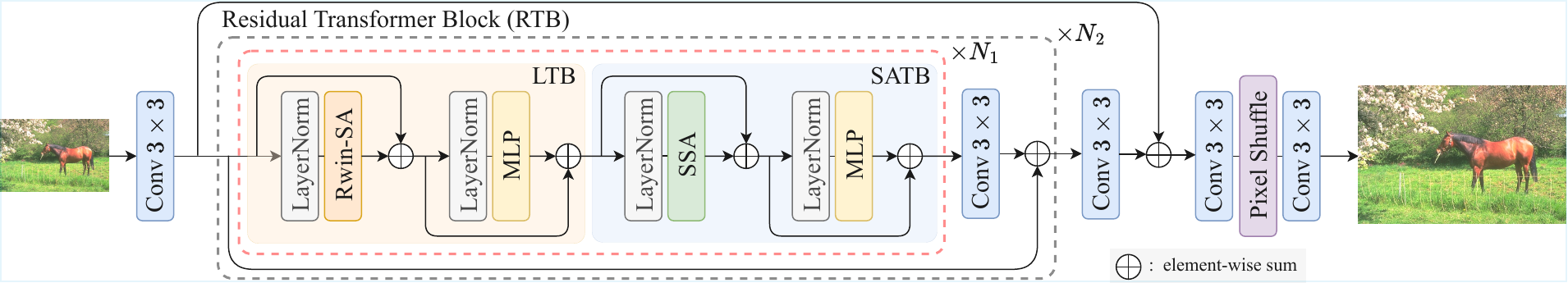}
    \caption{The architecture of the proposed SAT. The Local Transformer Block (LTB) and the Selective Aggregation Transformer Block (SATB) are arranged alternately to construct global-local structure, better capturing deep features for pixel reconstruction.}
    \label{fig:SAT}
    \vspace{-4mm}
\end{figure*}

\noindent \textbf{Token Reduction and Clustering Methods.} Token reduction methods aim to mitigate the quadratic complexity of vision transformers. DynamicViT \cite{rao2021dynamicvit}, Evo-ViT \cite{xu2022evo} progressively discard tokens based on token importance scores, but they sacrifice spatial information. ToMe \cite{bolya2022token} merges similar tokens using bipartite soft matching that limited to pairwise similarity and two-token merges at a time. DPC-KNN \cite{haurum2023tokens} adapts density-peak clustering \cite{rodriguez2014clustering} to ViTs to create semantical clusters to compress features. Overall, these methods share three key limitations for SR and other dense prediction tasks: (i) symmetric compression uniformly reducing query, key, and value that is suitable for classification but incompatible with SR requiring per-pixel predictions; (ii) density-based methods like DPC-KNN incur $\mathcal{O}(N^2)$ pairwise similarity computations that is impractical for online attention; (iii) uniform averaging in aggregation weakens feature norms, causing distributional shifts destabilize training. Our SAT mitigates these gaps via asymmetric Query-KeyValue aggregation, reducing center selection complexity to $\mathcal{O}(K^2)$, preserving feature norm distribution and dynamic integration within transformer architectures.

\section{Methodology}

\subsection{Motivation}

Vanilla self-attention is impractical for SR tasks due to its quadratic computational complexity, highlighting the need for an efficient approach that captures global dependencies at a low computational cost.
To this end, we analyze pixel-wise absolute error between SR outputs and GT images, we observe that the reconstruction error is concentrated in high-frequency regions (e.g., edges, textures), as in Fig. \ref{fig:motivation}. Even PFT achieves high performance, but is still struggling with these regions.
Our insight is that, in SR tasks, not all spatial locations contribute equally to reconstruction. Dense feature/high-frequency regions carry more information than homogeneous/low-frequency regions (e.g., smooth areas). Dense feature regions require global context to capture long-range dependencies, whereas low-frequency regions can be aggregated safely with minimal information loss. This imbalance motivates our Selective Aggregation Attention, which 
selectively merges low-frequency tokens for key-value projections during attention calculation, while preserving high-frequency tokens and maintaining critical details in query projection for high-quality reconstruction.

\subsection{Overall Framwork}

\begin{figure*}[t]
    \vspace{-2mm}
    \centering
    \includegraphics[width=0.95\textwidth]{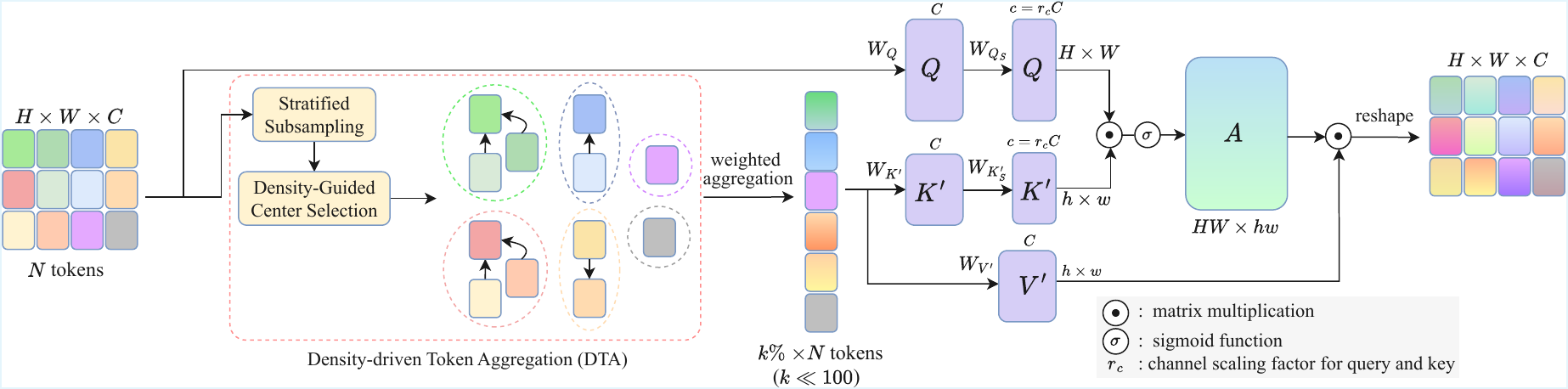}
    \vspace{-2mm}
    \caption{The illustration of the Selective Aggregation Attention (SAA). SAA aggregates $N$ input tokens into $K =  k\%\times N$ tokens (with $k = 3$) to compact the Key-Value matrices, preserving the full-resolution Query matrix to form an efficiently global cross-attention.}
    \label{fig:SAA}
    \vspace{-4mm}
\end{figure*}

The SAT's architecture is shown in Fig. \ref{fig:SAT}. 
SAT employs residual in residual structure to construct a deep feature extraction.
First, input image $I_{\text{LR}} \in \mathbb{R}^{H \times W \times 3}$ is embedded to $X_0 \in \mathbb{R}^{H \times W \times C}$ by a convolution layer. $H, W,$ $C$ are the image height, width, and channel count. $X_0$ is fed into the residual groups that include $N_2$ Residual Transformer Blocks (RTBs) to extract deep features, then passes it through a convolution to get refined features $X_1 \in \mathbb{R}^{H \times W \times C}$. Finally, $X_0$ and $X_1$ are fused via a residual connection 
and passed it into the upscaling module to get output image $I_{\text{SR}} \in \mathbb{R}^{sH \times sW \times C}$, where s is upscaling factor. 

Each RTB contains $N_1$ transformer blocks and a convolution. 
We use two types of transformer blocks: Local Transformer Blocks (LTB) and Selective Aggregation Transformer Blocks (SATB). These blocks are arranged in an alternating manner to establish a global-local structure. Our SATB focuses on global modeling while LTB assists in extracting local details that complement the deep feature extraction. Each block includes layer normalization, an a attention module, and a multilayer perceptron (MLP) \cite{vaswani2017attention}.

\subsection{Selective Aggregation Attention}

We formalize our Selective Aggregation Attention (SAA). Given an input feature $\mathbf{F} \in \mathbb{R}^{H \times W \times C}$, we first reshape into a token sequence $\mathbf{X} \in \mathbb{R}^{N \times C}$ where $N = HW$ is the sequence of tokens. Vanilla self-attention computes query, key, and value projections and attention output as:
{\fontsize{9pt}{12pt}\selectfont
\begin{subequations}
\begin{align}
\begin{split}
\mathbf{Q} = \mathbf{X}\mathbf{W}_Q, \mathbf{K} = \mathbf{X}\mathbf{W}_K, \mathbf{V} = \mathbf{X}\mathbf{W}_V, \label{e:vanilla1}
\vspace{-1mm}
 \end{split}\\
\begin{split}
\text{Attention}(\mathbf{Q}, \mathbf{K}, \mathbf{V}) = \text{softmax}(\frac{\mathbf{Q}\mathbf{K}^\top}{\sqrt{d}})\mathbf{V}, \label{e:vanilla2}
\vspace{-1mm}
\end{split}
\end{align}
\end{subequations}
}

\noindent where $\mathbf{W}_Q, \mathbf{W}_K, \mathbf{W}_V$ are learnable projections and $d$ is the attention head dimension.
\noindent Eq. \ref{e:vanilla2} requires $\mathcal{O}(N^2d)$ operations to compute the $N \times N$ matrix $\mathbf{Q}\mathbf{K}^\top$. In contrast, our SAA employs asymmetric compression, keeping \textit{full-resolution query} while \textit{compressing key and value} representations. We compute $\mathbf{Q} \in \mathbb{R}^{N \times d}$ as in vanilla self-attention, but use a selective aggregation operator $\Phi_{\text{SA}}: \mathbb{R}^{N \times d} \rightarrow \mathbb{R}^{K \times d}$ to $\mathbf{K}$ and $\mathbf{V}$, yielding $\mathbf{K}'$ and $\mathbf{V}' \in \mathbb{R}^{K \times d}$ as:
{\fontsize{9pt}{12pt}\selectfont
\begin{equation} \label{e:equation1}
    \mathbf{K}' = \Phi_{\text{SA}}(\mathbf{X}\mathbf{W}_K), \mathbf{V}' = \Phi_{\text{SA}}(\mathbf{X}\mathbf{W}_V),
    \vspace{-1mm}
\end{equation}
}

\noindent where $K$ is the number of compressed representations. To further reduce computations, we scale the channel dimension with scaling factor $r_c$ of $\mathbf{Q}$ and $\mathbf{K'}$ matrices through linear projections ($\mathbf{W}_{Q_S}, \mathbf{W}_{K_S^{'}}$), as shown in Fig. \ref{fig:SAA}. Then our SAA operates as cross-attention as:
\vspace{-1mm}
{\fontsize{9pt}{12pt}\selectfont
\begin{equation} \label{e:equation1}
    \text{SAA}(\mathbf{Q}, \mathbf{K}', \mathbf{V}') = \text{softmax}(\frac{\mathbf{Q}{\mathbf{K}'}^\top}{\sqrt{r_cd}})\mathbf{V}'
    \vspace{-1mm}
\end{equation}
}

\noindent This formulation reduces computational complexity from $\mathcal{O}(N^2d)$ to $\mathcal{O}(NKd)$ (we set $K \ll N$ to obtain much lower complexity) while preserving full spatial resolution in the output. 
By maintaining full-resolution query and compressing key and value, the design exploits the asymmetric information needs of SR: query preserves fine spatial structures for precise high-frequency detail recovery, whereas key and value can be compactly represented by prototype features. 
To better extract global-local contextual information, we combine our SAA with a recent local attention mechanism, Rwin-SA \cite{chen2022cross}, which is effective for diverse low-level vision tasks. Our ablations in Tab. \ref{tab:ab3} prove that our global-local structure design is an optimal choice for our network.

\subsection{Density-driven Token Aggregation}

We propose Density-driven Token Aggregation (DTA) as selective aggregation operator $\Phi_{\text{SA}}$. DTA is an efficient adaptation of density-peak clustering principles \cite{rodriguez2014clustering} specifically designed for high-dimensional vision token compression. $\Phi_{\text{SA}}$ takes $N$ input feature vectors and produces $K$ semantically representative vectors via the following steps: density-guided center selection with stratified subsampling, token assignment, and similarity-weighted aggregation.

\noindent \textbf{Density-Guided Center Selection.} Our DTA selects cluster centers with high local density, indicating many semantically similar neighbors, and large distances from other dense regions, ensuring clear inter-cluster boundaries. For each token $\mathbf{x}_i$, we compute its local density $\rho_i$ using a k-nearest neighbor estimator using cosine similarity as:
{\fontsize{9pt}{12pt}\selectfont
\vspace{-2mm}
\begin{subequations}
\begin{align}
\begin{split}
s(\mathbf{x}_i, \mathbf{x}_j) = \frac{\mathbf{x}_i^\top \mathbf{x}_j}{\|\mathbf{x}_i\| \|\mathbf{x}_j\|}, \vspace{-1mm} \label{e:cdg1}
 \end{split}\\
\begin{split}
\rho_i = m^{-1} \sum_{j \in \mathcal{N}_m(i)} s(\mathbf{x}_i, \mathbf{x}_j), 
\vspace{-1mm} \label{e:cdg2}
\end{split}
\vspace{-4mm}
\end{align}
\end{subequations}
}

\noindent where $\mathcal{N}_m(i)$ denotes $m$ nearest neighbors of token $i$.
We use cosine similarity instead of Euclidean distance, as angular relations better capture semantic similarity in high-dimensional visual feature spaces \cite{caron2021emerging, bolya2022token}, where magnitude-based distances suffer from concentration effects \cite{aggarwal2001surprising, beyer1999nearest}. 

The second quantity is the minimum distance to a higher density. We first convert cosine similarity to distance:
{\fontsize{9pt}{12pt}\selectfont
\begin{subequations}
\vspace{-1mm}
\begin{align}
\begin{split}
d(\mathbf{x}_i, \mathbf{x}_j) = 1 - s(\mathbf{x}_i, \mathbf{x}_j), \vspace{-4mm} \label{e:cdg1}
 \end{split}\\
\begin{split}
\delta_i = \min_{j: \rho_j > \rho_i} d(\mathbf{x}_i, \mathbf{x}_j), \label{e:cdg2}
\vspace{-2mm}
\end{split}
\end{align}
\end{subequations}
}


\noindent Typically, $\delta_i$ measures minimum distance 
to the nearest token with higher density $\rho_j > \rho_i$. For tokens 
at local density maxima, $\delta_i$ is 
set to the maximum distance to any token, ensuring these density peaks 
are prioritized as cluster centers.

The cluster-center selection criterion combines both properties into a unified score as:
{\fontsize{9pt}{12pt}\selectfont
\vspace{-1.5mm}
\begin{equation} \label{e:equation1}
    \gamma_i = \rho_i \cdot \delta_i,
    \vspace{-1.5mm}
\end{equation}
}

\noindent Tokens with high $\gamma$ values exhibit high local density and large separation (globally distinct), making them ideal cluster representatives. The $K$ highest-scoring tokens are selected as cluster centers $\mathcal{C} = \{\mathbf{c}_1, \ldots, \mathbf{c}_K\}$.

\noindent \textbf{Stratified Subsampling.} Computing density and separation measures across all $N$ tokens requires pairwise similarity evaluations, leading to $\mathcal{O}(N^2C)$ complexity that conflicts our efficiency objectives. To mitigate this while preserving representative feature coverage, we introduce a stratified subsampling strategy. Unlike naive random sampling that assumes tokens are independent and identically distributed, our method accounts for the spatial and semantic structure of natural images, where nearby pixels share similar features while distant regions often differ.

We first partition the $N$ tokens into $K$ spatially contiguous regions based on their raster-scan ordering in the feature map. Region boundaries are defined as follows:
{\fontsize{9pt}{12pt}\selectfont
\vspace{-0.7mm}
\begin{equation} \label{e:equation1}
    \mathcal{R}_i = \{j : (i-1)\lfloor \frac{N}{K} \rfloor \leq j < i\lfloor \frac{N}{K} \rfloor\}, \ i \in \{1, \ldots, K-1\}, \vspace{-0.5mm}
\end{equation}
}

\noindent The final region $\mathcal{R}_K$ contains all remaining tokens to handle non-divisibility. This partitioning maintains spatial continuity, ensuring each region forms a contiguous block in the feature map. From each $\mathcal{R}_i$, we uniformly sample $m_i = \lfloor \frac{S}{K} \rfloor$ tokens without replacement, where $S = \beta K$ is the target subsample size and $2 \leq \beta < \frac{N}{K}$ is the subsampling factor. Specifically, for each region, we compute:
$
m_i = \min\left(\left\lfloor \frac{S}{K} \right\rfloor, |\mathcal{R}_i|\right)
$
to avoid oversampling from regions containing fewer tokens than the target sample size. The regional subsamples $\mathcal{S}_i \subset \mathcal{R}_i$ with $|\mathcal{S}_i| = m_i$ are then merged to form the final subsample $\mathcal{S} = \bigcup_{i=1}^K \mathcal{S}_i$.
If the aggregate sample size $|\mathcal{S}| = \sum_{i=1}^K m_i$ is smaller than the target $S$ due to uneven region sizes or rounding, we augment $\mathcal{S}$ with additional tokens uniformly drawn from the remaining unsampled set.
With the subsample $\mathcal{S}$ constructed, we estimate density and separation statistics within this subset. The $S \times S$ subsampled similarity matrix $\mathbf{S}_{\mathcal{S}} = [s(\mathbf{x}_i, \mathbf{x}_j)]_{i,j \in \mathcal{S}}$ is formed, and for each token $i \in \mathcal{S}$, we obtain its local density $\tilde{\rho}_i$, separation $\tilde{\delta}_i$, and cluster-center score $\tilde{\gamma}_i = \tilde{\rho}_i \cdot \tilde{\delta}_i$. 
Top $K$ tokens with highest $\tilde{\gamma}_i$ values are selected as cluster centers and mapped back to their original indices in the full token sequence.


\noindent \textbf{Token Assignment and Similarity-Weighted Aggregation.} Following center selection, all $N$ tokens are assigned to their nearest cluster center based on cosine similarity:
{\fontsize{9pt}{12pt}\selectfont
\vspace{-1.0mm}
\begin{equation} \label{e:equation1}
    \alpha(i) = \text{argmax}_{k \in \{1,\ldots,K\}} s(\mathbf{x}_i, \mathbf{c}_k)
    \vspace{-1.0mm}
\end{equation}
}

\noindent 
Instead of uniform averaging that treats all cluster members equally regardless of their proximity to cluster center, we use similarity-weighted aggregation to merge tokens in each cluster while emphasizing semantically coherent members. For cluster $k$, the aggregated representation is computed as:
{\fontsize{9pt}{12pt}\selectfont
\vspace{-1.0mm}
\begin{equation} \label{e:equation1}
    \mathbf{y}_k = \frac{\sum_{i: \alpha(i)=k} w_i \mathbf{x}_i}{\sum_{i: \alpha(i)=k} w_i},
\end{equation}
}

\noindent where the weight $w_i = \exp(\frac{s(\mathbf{x}_i, \mathbf{c}_k)}{\tau})$ is based on the similarity between token $\mathbf{x}_i$ and center $\mathbf{c}_k$, scaled by temperature $\tau$. This design amplifies contributions from highly similar tokens while downweighting outliers. Temperature $\tau$ controls weighting sharpness: smaller values focus on close tokens, while larger values approximate uniform averaging.

However, weighted averaging systematically reduces feature magnitudes due to the triangle inequality:
{\fontsize{9pt}{12pt}\selectfont
\vspace{-1.0mm}
\begin{equation} \label{e:equation1}
    \|\sum_i w_i \mathbf{x}_i\| \leq \sum_i w_i \|\mathbf{x}_i\|, \vspace{-2.0mm}
\end{equation}
}

\noindent with equality only for parallel vectors. This norm reduction is problematic because feature magnitudes encode perceptually relevant information \cite{ho2020contrastive}, and layer normalization expects consistent magnitude distributions \cite{ba2016layer}. Therefore, we propose Feature Norm Restoration (FNR) as a post-processing step. Given original tokens $\{\mathbf{x}_1, \ldots, \mathbf{x}_N\}$ and weighted averages $\{\mathbf{y}_1, \ldots, \mathbf{y}_K\}$, we rescale them by global maximum norm as follows:
{\fontsize{9pt}{11pt}\selectfont
\vspace{-2mm}
\begin{subequations}
\begin{align}
\begin{split}
n_{\max} = \max_{i=1,\ldots,N} \|\mathbf{x}_i\|, \vspace{-1mm} \label{e:cdg1}
 \end{split}\\
\begin{split}
\mathbf{\hat{y}}_k = \begin{cases}
\frac{\mathbf{y}_k}{\|\mathbf{y}_k\|} \cdot n_{\max} & \text{if } \|\mathbf{y}_k\| > \epsilon \\
\mathbf{y}_k & \text{otherwise}
\end{cases} \label{e:cdg2} \vspace{-6mm}
\end{split}
\vspace{-6mm}
\end{align}
\end{subequations}
}
\noindent $\epsilon = 10^{-6}$ to avoid division by zero. This rescaling retains directional information from the weighted average and sets magnitude to the maximum observed in the original set, ensuring consistent feature statistics. We use global maximum instead of cluster-wise maxima to ensure uniform magnitude scaling over all $\mathbf{y}_{i}$, better keeping overall distribution. 

\subsection{Theoretical Analysis} 

We present a formal analysis of the complexity and approximation quality of our SAA. We believe that this theoretical analysis enhances the stability and reliability of SAA, providing a solid basis for interpreting our results.

\noindent \textbf{Theorem 3.1} (Computational Complexity). \textit{Our SAA reduces time complexity from $\mathcal{O}(N^2C)$ in vanilla self-attention to $\mathcal{O}(NKC)$, yielding a speedup factor of $\Theta(\frac{N}{K})$.}

\vspace{1mm}
\noindent \textbf{\textit{Proof.}} The computational cost of SAA includes the following parts: query projection $\mathcal{O}(NC^2)$; key and value projections each $\mathcal{O}(KC^2)$; Density-driven Token Aggregation $\mathcal{O}(NKC)$; computing attention matrix $\mathbf{Q}{\mathbf{K}'}^\top$ $\mathcal{O}(NKd)$; softmax $\mathcal{O}(NK)$; weighted aggregation $\mathbf{A}\mathbf{V}'$ $\mathcal{O}(NKd)$. The total complexity is $\mathcal{O}(NC^2 + K^2C + NKC + NKd)$. With $C > d$ and $K \ll N$ such that $K^2 \ll NK$, dominant term becomes $\mathcal{O}(NKC)$. Compared to vanilla self-attention's $\mathcal{O}(N^2C)$ yields speedup $\frac{\mathcal{O}(N^2C)}{\mathcal{O}(NKC)} = \Theta(\frac{N}{K})$. 

\noindent \textbf{Theorem 3.2} (Approximation Quality). \textit{Let $\mathbf{O}^* = \text{Attention}(\mathbf{Q}, \mathbf{K}, \mathbf{V})$ denote the vanilla self-attention output and $\mathbf{O} = \text{SAA}(\mathbf{Q}, \mathbf{K}', \mathbf{V}')$ denote the our SAA output. Under the assumptions that (i) the feature density field $\rho$ is Lipschitz continuous with constant $L$, (ii) features are sampled such that minimum inter-cluster separation exceeds $\epsilon > 0$, and (iii) subsampling size satisfies $S = \beta K$ with $\beta \geq 2$, given $\delta$ is a small failure probability parameter, the approximation error satisfies with probability at least $1 - \delta$:}
{\fontsize{9pt}{12pt}\selectfont
\begin{equation} \label{e:equation1}
    \|\mathbf{O} - \mathbf{O}^*\|_F \leq C_1 L\sqrt{\frac{NK\log(\delta ^{-1})}{S}} + C_2 \|\mathbf{V}\|_F \frac{K}{N}, \vspace{-1.mm}
\end{equation}
}

\noindent where $C_1, C_2$ are absolute constants, $\|\cdot\|_F$ is the Frobenius norm;
the first term captures clustering approximation error and the second term captures attention approximation error.

\vspace{1mm}
\noindent \textbf{\textit{Sketch Proof}}. We decomposes total error into two parts: clustering approximation and attention approximation. 

First, the \textbf{clustering approximation error} arises from using subsampled density estimates $\tilde{\rho}_i$ instead of exact densities $\rho_i$. By Hoeffding's inequality \cite{hoeffding1963probability}, each subsampled density estimate concentrates around its expectation with deviation 
$O(\sqrt{\log(\frac{\delta ^{-1})}{S}})$.
Under Lipschitz continuity of the density field, small perturbations in density estimates lead to controlled changes in the ranking induced by scores $\gamma_i = \rho_i \delta_i$. Aggregating over all $N$ tokens and $K$ clusters, and accounting for the assignment process, yields the first error term 
$O(L\sqrt{\frac{NK\log(\delta ^{-1})}{S}})$. 

Second, the \textbf{attention approximation error} stems from replacing the full $N \times N$ attention matrix with a compressed $N \times K$ cross-attention matrix. Each query's attention distribution over $K$ compressed keys approximates its distribution over the full $N$ keys by concentrating probability mass on cluster representatives. The quality of this approximation depends on within-cluster coherence, which is controlled by the clustering quality. Standard results on attention approximation combined with properties of the softmax function yield the second term $O(\|\mathbf{V}\|_F \frac{K}{N})$, capturing the relative error introduced by key compression. The final bound follows from the triangle inequality applied to these two components. \textit{The full proof is provided in the \textit{supp.} file.}


\section{Experiments}

\begin{table*}[t]
\centering
\setlength{\tabcolsep}{2.2mm}
{\fontsize{8pt}{8pt}\selectfont
\caption{Comparison between SAT and other SOTA methods at $\times$2, $\times$3, $\times$4 scales for image ISR. The top-2 results are in \textcolor{red}{red} and \textcolor{blue}{blue}.}
\vspace{-2mm}
\label{tab:table1}
\begin{tabular}{|l|c|c|c|c|c|c|c|c|c|c|c|c|c|}
\hline
\multirow{2}{*}{Method} & \multirow{2}{*}{Scale} & \multirow{2}{*}{Params} & \multirow{2}{*}{FLOPs} & \multicolumn{2}{c|}{Set5} & \multicolumn{2}{c|}{Set14} & \multicolumn{2}{c|}{B100} & \multicolumn{2}{c|}{Urban100} & \multicolumn{2}{c|}{Manga109} \\
\cline{5-14}
& & & & PSNR & \multicolumn{1}{c|}{SSIM} & PSNR & \multicolumn{1}{c|}{SSIM} & PSNR & \multicolumn{1}{c|}{SSIM} & PSNR & \multicolumn{1}{c|}{SSIM} & PSNR & SSIM \\
\hline
EDSR \cite{lim2017enhanced} & \multirow{10}{*}{$\times$2} &  42.6M  &  22.14T   &   38.11  & \multicolumn{1}{c|}{0.9692} & 33.92   & \multicolumn{1}{c|}{0.9195} &   32.32  & \multicolumn{1}{c|}{0.9013} &  32.93   & \multicolumn{1}{c|}{0.9351} &   39.10   &    0.9773 \\
RCAN \cite{zhang2018image} & &  15.4M  &  7.02T   &  38.27   & \multicolumn{1}{c|}{0.9614} &  34.12  & \multicolumn{1}{c|}{0.9216} &   32.41  & \multicolumn{1}{c|}{ 0.9027} &  33.34   & \multicolumn{1}{c|}{0.9384} &   39.44   &     0.9786 \\
IPT \cite{chen2021pre} & &  115M  &   7.38T  &  38.37   & \multicolumn{1}{c|}{-} &  34.43  & \multicolumn{1}{c|}{-} &  32.48   & \multicolumn{1}{c|}{-} &  33.76   & \multicolumn{1}{c|}{-} &   -   &    - \\
SwinIR \cite{liang2021swinir} & &  11.8M  &  3.04T   &  38.42   & \multicolumn{1}{c|}{0.9623} &  34.46  & \multicolumn{1}{c|}{ 0.9250} &   32.53  & \multicolumn{1}{c|}{0.9041} &  33.81   & \multicolumn{1}{c|}{0.9433} &  39.92    &    0.9797 \\
CAT-A \cite{chen2022cross} & &  16.5  &   5.08  &   38.51  & \multicolumn{1}{c|}{0.9626} &  34.78  & \multicolumn{1}{c|}{0.9265} &  32.59   & \multicolumn{1}{c|}{0.9047} &   34.26  & \multicolumn{1}{c|}{0.9440} &  40.10    &    0.9805 \\
HAT \cite{chen2023activating} & &  20.6M  &   5.81T  &  38.63   & \multicolumn{1}{c|}{0.9630} &  34.86  & \multicolumn{1}{c|}{0.9274} &   32.62  & \multicolumn{1}{c|}{0.9053} &  34.45   & \multicolumn{1}{c|}{0.9466} &   40.26   &   0.9809 \\
IPG \cite{tian2024image} & &  18.1M  &  5.35T   &  38.61   & \multicolumn{1}{c|}{0.9632} &  34.73  & \multicolumn{1}{c|}{0.9270} &   32.60  & \multicolumn{1}{c|}{0.9052} &  34.48   & \multicolumn{1}{c|}{0.9464} &  40.24    &   0.9810 \\
ATD \cite{zhang2024transcending} & &  20.1M  &  6.07T   &   38.61  & \multicolumn{1}{c|}{0.9629} &  34.95  & \multicolumn{1}{c|}{0.9276} &  32.65   & \multicolumn{1}{c|}{0.9056} &   34.70  & \multicolumn{1}{c|}{0.9476} &   40.37   &   0.9810 \\
PFT \cite{long2025progressive} & &  19.6M  &  5.03T   &  \textcolor{blue}{38.68}   & \multicolumn{1}{c|}{\textcolor{blue}{0.9635}} &  \textcolor{blue}{35.00}  & \multicolumn{1}{c|}{\textcolor{blue}{0.9280}} &  \textcolor{blue}{32.67}   & \multicolumn{1}{c|}{\textcolor{blue}{0.9058}} &   \textcolor{blue}{34.90}  & \multicolumn{1}{c|}{\textcolor{blue}{0.9490}} &   \textcolor{blue}{40.49}   &   \textcolor{blue}{0.9815} \\
\hdashline
\rowcolor{gray!13}
SAT (Ours) & &  19.4M  &  3.64T   &   \textcolor{red}{38.74}  & \multicolumn{1}{c|}{\textcolor{red}{0.9638}} &  \textcolor{red}{35.07}  & \multicolumn{1}{c|}{\textcolor{red}{0.9286}} &  \textcolor{red}{32.71}   & \multicolumn{1}{c|}{\textcolor{red}{0.9065}} &  \textcolor{red}{34.92}   & \multicolumn{1}{c|}{\textcolor{red}{0.9492}} &   \textcolor{red}{40.70}   &  \textcolor{red}{0.9818} \\
\hline

EDSR \cite{lim2017enhanced} & \multirow{10}{*}{$\times$3} &  43.0M  &  9.82T   &  34.65   & \multicolumn{1}{c|}{0.9280} &  30.52  & \multicolumn{1}{c|}{0.8462} &   29.25  & \multicolumn{1}{c|}{0.8093} &  28.80   & \multicolumn{1}{c|}{0.8653} &   34.17   &  0.9476 \\
RCAN \cite{zhang2018image} & &  15.6M  &   3.12T  &   34.74  & \multicolumn{1}{c|}{0.9299} &  30.65  & \multicolumn{1}{c|}{0.8482} &  29.32   & \multicolumn{1}{c|}{0.8111} &   29.09  & \multicolumn{1}{c|}{0.8702} &   34.44   &  0.9499 \\
IPT \cite{chen2021pre} & &  116M  &   3.28T  &   34.81  & \multicolumn{1}{c|}{-} &  30.85  & \multicolumn{1}{c|}{-} &   29.38  & \multicolumn{1}{c|}{-} &   29.49  & \multicolumn{1}{c|}{-} &   -   &   - \\
SwinIR \cite{liang2021swinir} & &  11.9M  &  1.35T   &  34.97   & \multicolumn{1}{c|}{ 0.9318} &  30.93  & \multicolumn{1}{c|}{ 0.8534} &  29.46   & \multicolumn{1}{c|}{0.8145} &   29.75  & \multicolumn{1}{c|}{0.8826} &   35.12   &   0.9537 \\
CAT-A \cite{chen2022cross} & &  16.6M  &   2.26T  &  35.06   & \multicolumn{1}{c|}{0.9326} &  31.04  & \multicolumn{1}{c|}{0.8538} &  29.52   & \multicolumn{1}{c|}{0.8160} &  30.12   & \multicolumn{1}{c|}{0.8862} &  35.38    &  0.9546 \\
HAT \cite{chen2023activating} & &  20.8M  &   2.58T  &   35.07  & \multicolumn{1}{c|}{ 0.9329} &  31.08  & \multicolumn{1}{c|}{0.8555} &  29.54   & \multicolumn{1}{c|}{0.8167} &   30.23  & \multicolumn{1}{c|}{ 0.8896} &   35.53   &    0.9552 \\
IPG \cite{tian2024image} & &  18.3M  &  2.39T   &   35.10  & \multicolumn{1}{c|}{0.9332} &  31.10  & \multicolumn{1}{c|}{0.8554} &  29.53   & \multicolumn{1}{c|}{0.8168} &   30.36  & \multicolumn{1}{c|}{0.8901} &   35.53   &  0.9554 \\
ATD \cite{zhang2024transcending} & &  20.3M  &  2.69T   &  35.11   & \multicolumn{1}{c|}{0.9330} &  31.13  & \multicolumn{1}{c|}{0.8556} &  29.57   & \multicolumn{1}{c|}{0.8176} &  30.46   & \multicolumn{1}{c|}{0.8917} &   35.63   &  0.9558 \\
PFT \cite{long2025progressive} & &  19.8M  &  2.23T   &  \textcolor{blue}{35.15}  & \multicolumn{1}{c|}{\textcolor{blue}{0.9333}} &  \textcolor{blue}{31.16}  & \multicolumn{1}{c|}{\textcolor{blue}{0.8561}} &   \textcolor{blue}{29.58}  & \multicolumn{1}{c|}{\textcolor{blue}{0.8178}} &   \textcolor{blue}{30.56}  & \multicolumn{1}{c|}{\textcolor{blue}{0.8931}} &   \textcolor{blue}{35.67}   &  \textcolor{blue}{0.9560} \\
\hdashline
\rowcolor{gray!13}
SAT (Ours) & &  19.5M  &   1.63T  &   \textcolor{red}{35.26}  & \multicolumn{1}{c|}{\textcolor{red}{0.9341}} & \textcolor{red}{31.22}   & \multicolumn{1}{c|}{\textcolor{red}{0.8569}} &  \textcolor{red}{29.63}   & \multicolumn{1}{c|}{\textcolor{red}{0.8186}} &  \textcolor{red}{30.67}   & \multicolumn{1}{c|}{\textcolor{red}{0.8949}} &   \textcolor{red}{35.87}   &  \textcolor{red}{0.9568} \\
\hline

EDSR \cite{lim2017enhanced} & \multirow{10}{*}{$\times$4} &  43.0M  &  5.54T   &  32.46   & \multicolumn{1}{c|}{0.8968} &  28.80  & \multicolumn{1}{c|}{0.7876} &  27.71   & \multicolumn{1}{c|}{0.7420} &   26.64  & \multicolumn{1}{c|}{0.8033} &   31.02   &   0.9148 \\
RCAN \cite{zhang2018image} & &  15.6M  &   1.76T  &  32.63   & \multicolumn{1}{c|}{0.9002} &  28.87  & \multicolumn{1}{c|}{ 0.7889} &  27.77   & \multicolumn{1}{c|}{0.7436} &   26.82  & \multicolumn{1}{c|}{0.8087} &  31.22    &  0.9173 \\
IPT \cite{chen2021pre} & &  116M  &  1.85T   &   32.64  & \multicolumn{1}{c|}{-} &  29.01  & \multicolumn{1}{c|}{-} &   27.82  & \multicolumn{1}{c|}{-} &   27.26  & \multicolumn{1}{c|}{-} &   -   &    - \\
SwinIR \cite{liang2021swinir} & &  11.9M  &  0.76T   &   32.92  & \multicolumn{1}{c|}{ 0.9044} &  29.09  & \multicolumn{1}{c|}{0.7950} &  27.92   & \multicolumn{1}{c|}{0.7489} &  27.45   & \multicolumn{1}{c|}{0.8254} &   32.03   &   0.9260 \\
CAT-A \cite{chen2022cross} & &  16.6M  &  1.27T   &   33.08  & \multicolumn{1}{c|}{ 0.9052} &  29.18  & \multicolumn{1}{c|}{0.7960} &  27.99   & \multicolumn{1}{c|}{0.7510} &  27.89   & \multicolumn{1}{c|}{0.8339} &   32.39   &    0.9285 \\
HAT \cite{chen2023activating} & &  20.8M  &  1.45T   &   33.04  & \multicolumn{1}{c|}{ 0.9056} &  29.23  & \multicolumn{1}{c|}{0.7973} &  28.00   & \multicolumn{1}{c|}{0.7517} &   27.97  & \multicolumn{1}{c|}{ 0.8368} &   32.48   & 0.9292 \\
IPG \cite{tian2024image} & &  18.3M  &   1.30T  &  \textcolor{blue}{33.15}   & \multicolumn{1}{c|}{0.9062} &  29.24  & \multicolumn{1}{c|}{0.7973} &   27.99  & \multicolumn{1}{c|}{0.7519} &   28.13  & \multicolumn{1}{c|}{ 0.8392} &   32.53   &  0.9300 \\
ATD \cite{zhang2024transcending} & &  20.3M  &  1.52T   &   33.10  & \multicolumn{1}{c|}{0.9058} &  29.24  & \multicolumn{1}{c|}{0.7974} &  28.01   & \multicolumn{1}{c|}{0.7526} &  28.17   & \multicolumn{1}{c|}{0.8404} &   32.62   &    \textcolor{blue}{0.9306}\\
PFT \cite{long2025progressive} & &  19.8M  &  1.26T   &  \textcolor{blue}{33.15}   & \multicolumn{1}{c|}{\textcolor{blue}{0.9065}} &  \textcolor{blue}{29.29}  & \multicolumn{1}{c|}{\textcolor{blue}{0.7978}} &   \textcolor{blue}{28.02}  & \multicolumn{1}{c|}{\textcolor{blue}{0.7527}} &  \textcolor{blue}{28.20}   & \multicolumn{1}{c|}{\textcolor{blue}{0.8412}} &   \textcolor{blue}{32.63}   &  \textcolor{blue}{0.9306} \\
\hdashline
\rowcolor{gray!13}
SAT (Ours) & &  19.5M  &  0.94T   &   \textcolor{red}{33.19}  & \multicolumn{1}{c|}{\textcolor{red}{0.9073}} & \textcolor{red}{29.35}   & \multicolumn{1}{c|}{\textcolor{red}{0.7996}} &  \textcolor{red}{28.08}   & \multicolumn{1}{c|}{\textcolor{red}{0.7535}} &  \textcolor{red}{28.29}   & \multicolumn{1}{c|}{\textcolor{red}{0.8423}} &  \textcolor{red}{32.85}   &   \textcolor{red}{0.9314} \\
\hline
\end{tabular}}
\vspace{-2mm}
\end{table*}

\subsection{Experimental Settings}
Following recent SR methods \cite{chen2023activating, chen2023recursive, tian2024image}, we use DFT2K (DIV2K \cite{lim2017enhanced} + Flicker2K \cite{timofte2017ntire}), a dataset widely used for ISR, as training dataset. For testing, we adopt five benchmark datasets: Set5 \cite{bevilacqua2012low}, Set14 \cite{zeyde2010single}, B100 \cite{arbelaez2010contour}, Urban100 \cite{huang2015single}, and Manga109 \cite{matsui2017sketch}. 
We evaluate our model's performance using the metrics PSNR and SSIM \cite{wang2004image}, calculated on the Y channel. The details of the training procedure and network hyperparameters can be found in the \textit{supp.} file.

\subsection{Comparisons with State-of-the-art Methods}

\textbf{Quantitative results.} Tab. \ref{tab:table1} presents PSNR and SSIM results, showing that our SAT outperforms all recent methods, including: EDSR \cite{lim2017enhanced}, RCAN \cite{zhang2018image}, IPT \cite{chen2021pre}, SwinIR \cite{liang2021swinir}, CAT-A \cite{chen2022cross}, HAT \cite{chen2023activating}, IPG \cite{tian2024image}, ATD \cite{zhang2024transcending} and PFT \cite{long2025progressive} across all three scales and various benchmarks. Notably, SAT surpasses the current SOTA method, PFT, while using fewer parameters and FLOPs. For instance, at $\times$4 scale, SAT achieves a maximum improvement of \textbf{0.22dB} on Manga109 compared to PFT, while reducing FLOPs by \textbf{25\%}, and it even reduces \textbf{27\%} FLOPs at $\times$2; showing a substantial improvement in image SR. All FLOPs in this paper are computed based on an HR image with a resolution of $1280\times 640$. 
We also compare our SAT-light (a small version of SAT, see \textit{supp.} file) with existing methods, as in Tab. \ref{tab:table2}, on lightweight benchmark to show its robustness and scalability. The results show that SAT-light consistently outperforms all methods while reducing \textbf{FLOPs by nearly half}, showing its efficiency. SAT’s superior performance stems from Selective Aggregation Attention, an asymmetric Query-KeyValue compression mechanism that efficiently models global dependencies. This enhances the reconstruction of high-frequency information by focusing on challenging regions while safely aggregating similar smooth areas, thereby significantly reducing computations.

\begin{table*}[t]
\centering
\setlength{\tabcolsep}{1.98mm}
{\fontsize{8pt}{8pt}\selectfont
\caption{Comparison between SAT-light and other methods at $\times$2, $\times$4 scales on lightweight benchmark. Top-2 results are in \textcolor{red}{red} and \textcolor{blue}{blue}.}
\vspace{-2mm}
\label{tab:table2}
\begin{tabular}{|l|c|c|c|c|c|c|c|c|c|c|c|c|c|}
\hline
\multirow{2}{*}{Method} & \multirow{2}{*}{Scale} & \multirow{2}{*}{Params} & \multirow{2}{*}{FLOPs} & \multicolumn{2}{c|}{Set5} & \multicolumn{2}{c|}{Set14} & \multicolumn{2}{c|}{B100} & \multicolumn{2}{c|}{Urban100} & \multicolumn{2}{c|}{Manga109} \\
\cline{5-14}
& & & & PSNR & \multicolumn{1}{c|}{SSIM} & PSNR & \multicolumn{1}{c|}{SSIM} & PSNR & \multicolumn{1}{c|}{SSIM} & PSNR & \multicolumn{1}{c|}{SSIM} & PSNR & SSIM \\
\hline
CARN \cite{ahn2018fast} & \multirow{10}{*}{$\times$2} &  1,592K  &  222.8G   &  37.76   & \multicolumn{1}{c|}{0.9590} & 33.52   & \multicolumn{1}{c|}{0.9166} &  32.09   & \multicolumn{1}{c|}{0.8978} &  31.92   & \multicolumn{1}{c|}{0.9256} &   38.36   &   0.9765 \\
IMDN \cite{hui2019lightweight} & &  694K  &   158.8G  &  38.00   & \multicolumn{1}{c|}{ 0.9605} &  33.63  & \multicolumn{1}{c|}{0.9177} &   32.19  & \multicolumn{1}{c|}{0.8996} &  32.17   & \multicolumn{1}{c|}{0.9283} &   38.88   &  0.9774 \\
LatticeNet \cite{luo2020latticenet} & &  756K  &  169.5G   &  38.15   & \multicolumn{1}{c|}{0.9610} &  33.78  & \multicolumn{1}{c|}{0.9193} &  32.25   & \multicolumn{1}{c|}{0.9005} &  32.43   & \multicolumn{1}{c|}{0.9302} &   -   &    - \\
SwinIR-light \cite{liang2021swinir} & &  910K  &  244G   &   38.14  & \multicolumn{1}{c|}{0.9611} &  33.86  & \multicolumn{1}{c|}{0.9206} &  32.31   & \multicolumn{1}{c|}{ 0.9012} &  32.76   & \multicolumn{1}{c|}{0.9340} &   39.12   &   0.9783 \\
ELAN \cite{zhang2022efficient} & &  582K  &  203G   &   38.17  & \multicolumn{1}{c|}{0.9611} &  33.94  & \multicolumn{1}{c|}{0.9207} &   32.30  & \multicolumn{1}{c|}{0.9012} &  32.76   & \multicolumn{1}{c|}{0.9340} &   39.11   &  0.9782 \\
OmniSR \cite{wang2023omni} & &  772K  &   194.5G  &  38.22   & \multicolumn{1}{c|}{0.9613} & 33.98   & \multicolumn{1}{c|}{0.9210} &   32.36  & \multicolumn{1}{c|}{0.9020} &  33.05   & \multicolumn{1}{c|}{ 0.9363} &   39.28   &   0.9784 \\
IPG-Tiny \cite{tian2024image} & &  872K  &  245.2G   &  38.27   & \multicolumn{1}{c|}{0.9616} &  \textcolor{blue}{34.24}  & \multicolumn{1}{c|}{\textcolor{blue}{0.9236}} &   32.35  & \multicolumn{1}{c|}{0.9018} &   33.04  & \multicolumn{1}{c|}{0.9359} &   39.31   &  0.9786 \\
ATD-light \cite{zhang2024transcending} & &  753K  &  348.6G   &  38.28   & \multicolumn{1}{c|}{0.9616} &  34.11  & \multicolumn{1}{c|}{ 0.9217} &   32.39  & \multicolumn{1}{c|}{0.9023} &  \textcolor{blue}{33.27}   & \multicolumn{1}{c|}{0.9376} &   39.51   &   0.9789 \\
PFT-light \cite{long2025progressive} & &  776K  &  278.3G   &   \textcolor{blue}{38.36}  & \multicolumn{1}{c|}{\textcolor{blue}{0.9620}} &  34.19  & \multicolumn{1}{c|}{0.9232} &  \textcolor{blue}{32.43}   & \multicolumn{1}{c|}{\textcolor{blue}{0.9030}} &  \textcolor{red}{33.67}   & \multicolumn{1}{c|}{\textcolor{red}{0.9411}} &   \textcolor{blue}{39.55}   &  \textcolor{blue}{0.9792} \\
\hdashline
\rowcolor{gray!13}
SAT-light (Ours) & &  742K  &  145.7G   &  \textcolor{red}{38.38}   & \multicolumn{1}{c|}{\textcolor{red}{0.9621}} &  \textcolor{red}{34.21}  & \multicolumn{1}{c|}{\textcolor{red}{0.9238}} &  \textcolor{red}{32.45}   & \multicolumn{1}{c|}{\textcolor{red}{0.9032}} &   \textcolor{red}{32.67}  & \multicolumn{1}{c|}{\textcolor{blue}{0.9410}} &   \textcolor{red}{39.71}   &   \textcolor{red}{0.9794} \\
\hline

CARN \cite{ahn2018fast} & \multirow{10}{*}{$\times$4} &  1,592K  &  90.9G   &  32.13   & \multicolumn{1}{c|}{0.8937} &  28.60  & \multicolumn{1}{c|}{0.7806} &  27.58   & \multicolumn{1}{c|}{0.7349} &  26.07   & \multicolumn{1}{c|}{0.7837} &   30.47   &  0.9084 \\
IMDN \cite{hui2019lightweight} & &  715K  &  40.9G   &   32.21  & \multicolumn{1}{c|}{0.8948} &  28.58  & \multicolumn{1}{c|}{0.7811} &  27.56   & \multicolumn{1}{c|}{0.7353} & 26.04    & \multicolumn{1}{c|}{0.7838} &   30.45   &   0.9075 \\
LatticeNet \cite{luo2020latticenet} & &  777K  &  43.6G   &   32.30  & \multicolumn{1}{c|}{0.8962} &  28.68  & \multicolumn{1}{c|}{0.7830} &  27.62   & \multicolumn{1}{c|}{0.7367} &  26.25   & \multicolumn{1}{c|}{0.7873} &   -   &    - \\
SwinIR-light \cite{liang2021swinir} & &  930K  &  63.6G   &  32.44   & \multicolumn{1}{c|}{0.8976} &  28.77  & \multicolumn{1}{c|}{0.7858} &  27.69   & \multicolumn{1}{c|}{0.7406} &   26.47  & \multicolumn{1}{c|}{0.7980} &   30.92   &  0.9151 \\
ELAN \cite{zhang2022efficient} & &  582K  &  54.1G   &  32.43   & \multicolumn{1}{c|}{0.8975} &  28.78  & \multicolumn{1}{c|}{0.7858} &  27.69   & \multicolumn{1}{c|}{0.7406} &  26.54   & \multicolumn{1}{c|}{0.7982} &   30.92   &   0.9150 \\
OmniSR \cite{wang2023omni} & &  792K  &  50.9G   &  32.49   & \multicolumn{1}{c|}{0.8988} &  28.78  & \multicolumn{1}{c|}{0.7859} &  27.71   & \multicolumn{1}{c|}{0.7415} &  26.65   & \multicolumn{1}{c|}{ 0.8018} &   31.02   &   0.9151 \\
IPG-Tiny \cite{tian2024image} & &  887K  &  61.3G   &  32.51   & \multicolumn{1}{c|}{0.8987} &  28.85  & \multicolumn{1}{c|}{ 0.7873} &  27.73   & \multicolumn{1}{c|}{0.7418} &  26.78   & \multicolumn{1}{c|}{ 0.8050} &   31.22   &   0.9176\\
ATD-light \cite{zhang2024transcending} & &  769K  &  87.1G   &   32.62  & \multicolumn{1}{c|}{0.8997} &  28.87  & \multicolumn{1}{c|}{0.7884} &   27.77  & \multicolumn{1}{c|}{0.7439} &  26.97   & \multicolumn{1}{c|}{0.8107} &   31.47   &  0.9198 \\
PFT-light \cite{long2025progressive} & &  792K  &  69.6G   &  \textcolor{blue}{32.63}   & \multicolumn{1}{c|}{\textcolor{blue}{0.9005}} &  \textcolor{blue}{28.92}  & \multicolumn{1}{c|}{\textcolor{blue}{0.7891}} &  \textcolor{blue}{27.79}   & \multicolumn{1}{c|}{\textcolor{blue}{0.7445}} &  \textcolor{blue}{27.20}   & \multicolumn{1}{c|}{\textcolor{blue}{0.8171}} &   \textcolor{blue}{31.51}   & \textcolor{blue}{0.9204} \\
\hdashline
\rowcolor{gray!13}
SAT-light (Ours) & &  763K  &   36.4G  &   \textcolor{red}{32.67}  & \multicolumn{1}{c|}{\textcolor{red}{0.9006}} &  \textcolor{red}{28.98}  & \multicolumn{1}{c|}{\textcolor{red}{0.7894}} &  \textcolor{red}{27.83}   & \multicolumn{1}{c|}{\textcolor{red}{0.7449}} &  \textcolor{red}{27.22}   & \multicolumn{1}{c|}{\textcolor{red}{0.8172}} &   \textcolor{red}{31.66}   &  \textcolor{red}{0.9205} \\
\hline
\end{tabular}}
\end{table*}

\begin{figure*}[t]
    \centering
    \includegraphics[width=0.9\textwidth]{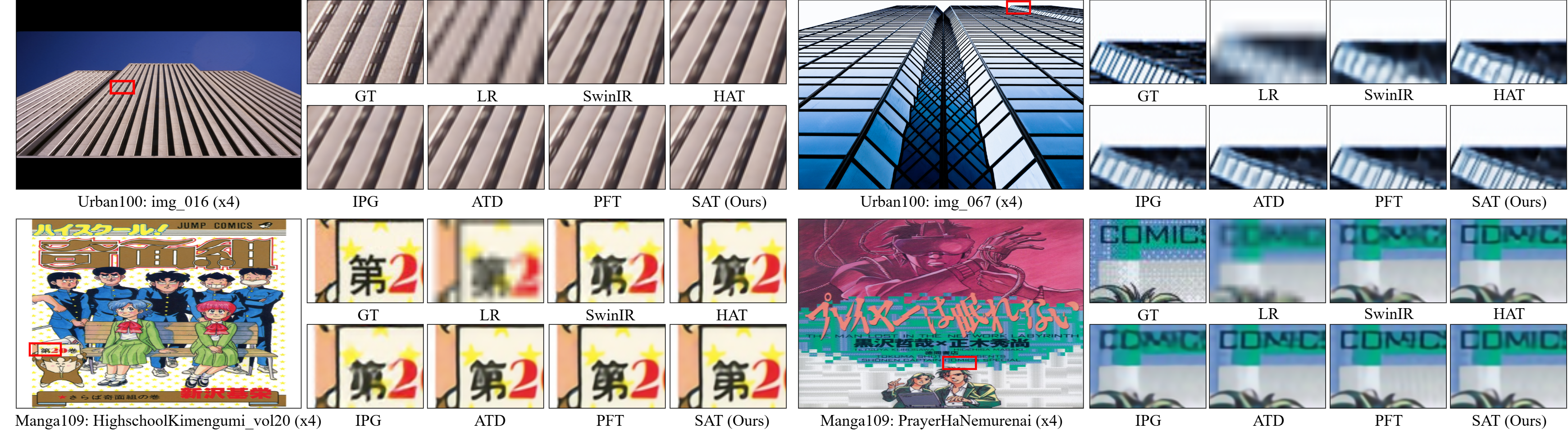}
    \vspace{-4mm}
    \caption{Qualitative comparison of visual results between our SAT and other state-of-the-art SR methods. Best results are marked in \textbf{bold}.}
    \label{fig:visual_result}
    \vspace{3mm}
    \includegraphics[width=0.9\textwidth]{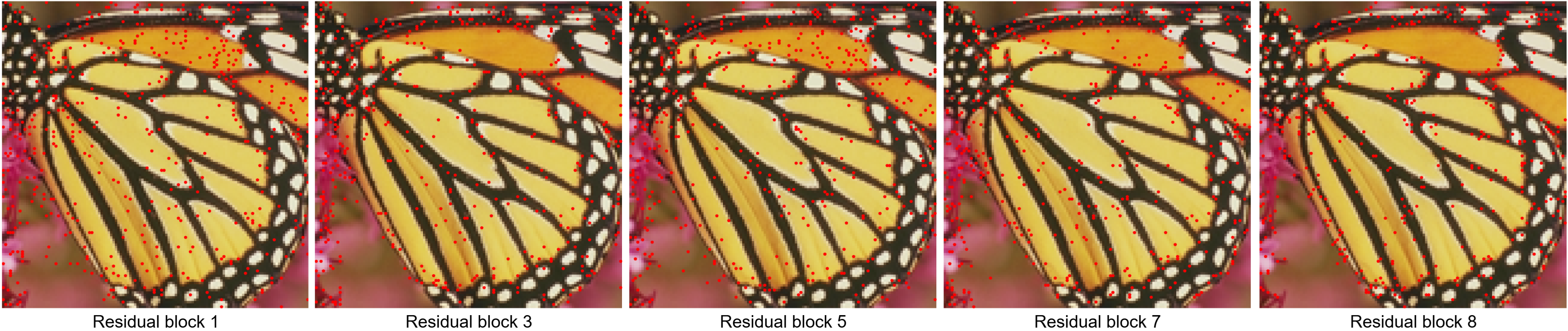}
    \vspace{-4mm}
    \caption{Visualization on low-resolution input for cluster center selection (red points) across different network layers. Early layers maintain broad spatial coverage, while deeper layers increasingly concentrate on semantically salient regions such as edges and pattern features. This progressive adaptation enables efficient compression without exhaustive spatial coverage. 
    }
    \label{fig:center_selection}
    \vspace{-1mm}
\end{figure*}

\noindent \textbf{Visual comparison.} We present visual results of various methods in Fig. \ref{fig:visual_result}. As illustrated, our SAT method is better in producing edges or textual detail while generating fewer artifacts 
compared to other approaches. In contrast, the other approaches cannot restore correct textures or hallucinate fine-grained details.
We also visualize cluster center selection on a low-resolution input across different SAA layers, specifically the final SAA layers of Residual Blocks 1, 3, 5, 7, and 8. Early layers (Block 1) maintain broad spatial coverage, while deeper layers (Block 7-8) increasingly concentrate on semantically salient regions such as edges and pattern features. This progressive adaptation shows the content-aware nature of our DTA algorithm, enabling efficient compression without exhaustive spatial coverage. The visualization shows centers capture sufficient diversity for attention to work well.  Note that our method is not designed to find optimal semantic clusters; we prioritize efficient attention approximation quality, achieving substantial speedup (Theorem 3.1) while maintaining reconstruction fidelity. More qualitative results can be found in \textit{supp.} file.

\subsection{Ablation Study}

\begin{figure}[t]
    \vspace{-2.5mm}
    \centering
    \includegraphics[width=0.47\textwidth]{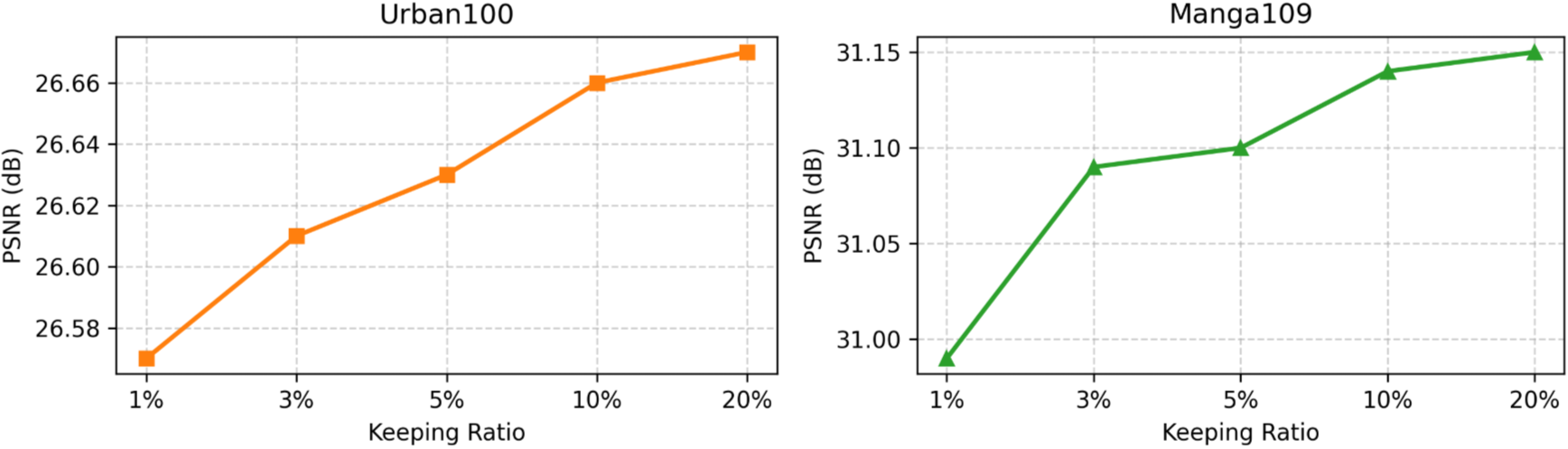}
    \vspace{-2mm}
    \caption{PSNR performance across different keeping ratios on Urban100, and Manga109 datasets.}
    \label{fig:keeping_ratio}
    \vspace{-2mm}
\end{figure}

We conduct extensive ablations to understand our proposal better. Following \cite{long2025progressive}, we perform all experiments at $\times 4$ scale for 250k iters on DIV2K with batch size 8. Due to page limit, more ablations are put in the \textit{supp.} file.

\noindent \textbf{Effects of Selective Aggregation Attention.} Tab. \ref{tab:ab1} shows the effectiveness of our Selective Aggregation Attention (SAA) compared to vanilla self-attention (VSA) \cite{vaswani2017attention}, spatial-reduction attention (SRA) from PVT \cite{wang2021pyramid}, and window self-attention (WSA) \cite{chen2022cross}. VSA achieves the best performance; however, it consumes significantly more FLOPs and, especially, VRAM, dominating other methods. Our SAA provides a better trade-off between complexity and performance, achieving performance close to VSA while requiring much less FLOPs and VRAM. Compared to SRA from PVT and WSA, our method shows similar FLOPs and VRAM consumption but delivers superior performance.

\begin{table}[h]
\centering
\setlength{\tabcolsep}{1.mm}
{\fontsize{8pt}{9.5pt}\selectfont
\caption{Effects of the proposed selective aggregation attention}
\vspace{-2mm}
\label{tab:ab1}
\begin{tabular}{lcccccccccc}
\toprule
Method & Params & FLOPs & VRAM & Set5 & Urban100 & Manga109 \\
\midrule
VSA \cite{vaswani2017attention} & 808K & 69.4G & 60.4GB & \textbf{32.48} & \textbf{26.74} & \textbf{31.12} \\
SRA \cite{wang2021pyramid} & 787K & 37.5G & 4.7GB & 32.40 & 26.47 & 30.88 \\
WSA \cite{liang2021swinir} & 809K & 43.9G & 4.1GB & \underline{32.44} & 26.52 & 30.92 \\
SAA (Ours) & 763K &  36.4G & 5.3GB & \textbf{32.48} & \underline{26.61} & \underline{31.09} \\
\bottomrule
\end{tabular}}
\vspace{-2mm}
\end{table}

\noindent \textbf{Effects of Density-driven Token Aggregation.}
Tab. \ref{tab:ab2} compares our DTA with two common clustering algorithms, K-means \cite{mcqueen1967some} (20 iterations) and DPC-KNN \cite{rodriguez2014clustering}. As shown, our method achieves the lowest time complexity, whereas DPC-KNN suffers from quadratic complexity, resulting in extremely long runtimes that make it impractical for training SR model. Compared to K-Means, our approach runs 10$\times$ faster in this ablation. In terms of performance, DTA achieves results comparable to DPC-KNN while significantly reducing runtime, demonstrating the robustness and efficiency of the proposed method. Without DTA, our SAA become VSA as in Tab. \ref{tab:ab1}.

\begin{table}[h]
\centering
\setlength{\tabcolsep}{0.3mm}
{\fontsize{8pt}{9.5pt}\selectfont
\caption{Effects of Density-driven Token Aggregation algorithm}
\vspace{-2mm}
\label{tab:ab2}
\begin{tabular}{lcccccccccc}
\toprule
Method & Complexity & Runtime & Set5 & Urban100 & Manga109 \\
\midrule
K-means (20 iters) \cite{mcqueen1967some} & $\mathcal{O}(20NKC)$ & \underline{113ms} & 32.39 & 26.49 & 30.91 \\
DPC-KNN \cite{rodriguez2014clustering} & $\mathcal{O}(N^{2}C)$ & 6534ms & \textbf{32.50} & \textbf{26.66} & \textbf{31.14} \\
DTA (Ours) & $\mathcal{O}(NKC)$ &  \textbf{11ms} & \underline{32.48} & \underline{26.61} & \underline{31.09} \\
\bottomrule
\end{tabular}}
\end{table}

\noindent \textbf{Effects of Compression Level.} 
Fig. \ref{fig:keeping_ratio} shows the trade-off between the token keeping ratio and PSNR performance. It shows that even with a small keeping ratio, it has small impact on reconstruction quality. PSNR steadily increases as the keeping ratio rises from 1\% to 20\%, but the improvement slows notably beyond 10\%, indicating that performance saturates and cannot be enhanced by merely increasing the keeping ratio. A larger keeping ratio pushes SAT closer to the complexity of vanilla self-attention. We select a keeping ratio of 3\% (removing 97\% of the tokens in the Key and Value matrices) as the final choice to balance super-resolution quality and model complexity.


\noindent \textbf{Effects of Global-Local Transformer Design.} 
The experiments in Tab. \ref{tab:ab3} verifies that our global–local hybrid design is an optimal choice for SR models. We sequentially remove the LTB and SATB modules in the first and second rows, respectively, while the last row uses an alternating configuration of LTB and SATB. The results show that using only our SATB already yields strong performance, and adding LTB further improves PSNR. Therefore, we adopt this design as our final architecture to achieve SOTA performance with manageable computational cost.
\begin{table}[h]
\centering
\setlength{\tabcolsep}{1.3mm}
{\fontsize{9pt}{12pt}\selectfont
\caption{Effects of SATB and LTB blocks.}
\vspace{-2mm}
\label{tab:ab3}
\begin{tabular}{ccccccc}
\toprule
LTB & SATB & Params & FLOPs & Set5 & Urban100 & Manga 109 \\
\midrule
w/o & w/ & 716K & 28.6G & \underline{32.45} & \underline{26.58} & \underline{31.07} \\
w/ & w/o & 810K & 44.2G & 32.41 & 26.48 & 30.97 \\
w/ & w/ & 763K & 36.4G & \textbf{32.48} & \textbf{26.61} & \textbf{31.09} \\
\bottomrule
\end{tabular}}
\vspace{-2mm}
\end{table}

\subsection{Model Complexity and Runtime Analysis}

We compare the complexity and inference time of our SAT with several SOTA methods, including HAT \cite{chen2023activating}, IPG \cite{tian2024image}, ATD \cite{zhang2024transcending}, and PFT \cite{long2025progressive}. In this experiment, the inference time for all models is measured on a NVIDIA RTX PRO 6000 GPU with 96GB of VRAM at an output resolution of $512 \times 512$. As shown in Tab. \ref{tab:runtime}, the inference time of our SAT is comparable to existing methods. Our model is a bit slower than HAT but is substantially better in term of performance. SAT also achieves lower computational complexity, and delivers the best reconstruction performance among current SOTA methods, including ATD, IPG and PFT. Comparison on $\times$ 2 and $\times$ 3 scales are reported in the \textit{supp.} file.

\begin{table}[h]
\centering
\setlength{\tabcolsep}{0.85mm}
\vspace{-2mm}
{\fontsize{8.7pt}{10pt}\selectfont
\caption{Comparison on model complexity and running time}
\vspace{-2mm}
\label{tab:runtime}
\begin{tabular}{clcccccccccc}
\toprule
Scale & Method & Params & FLOPs & PSNR (Manga109) & Runtime \\
\midrule
$\times$4 & HAT \cite{chen2023activating} & 20.8M & 1.45T & 32.48 & \textbf{192ms} \\
$\times$4 & ATD \cite{zhang2024transcending} & 20.3M & 1.52T & 32.62 & 228ms \\
$\times$4 & IPG \cite{tian2024image} & 18.3M & 1.30T & 32.53 & 288ms \\
$\times$4 & PFT \cite{long2025progressive} & 19.8M &  1.26T & \underline{32.63} & 230ms \\
$\times$4 &SAT (Ours) & 19.5M & 0.94T & \textbf{32.85} & \underline{207ms} \\
\bottomrule
\end{tabular}}
\end{table}

\vspace{-2mm}
\section{Conclusion}

In this study, we propose a novel Selective Aggregation Transformer, SAT, for image SR. 
The key component of SAT is Selective Aggregation Attention, which approximates global attention efficiently. Specifically, we employ an asymmetric Query-KeyValue compression through our Density-driven Token Aggregation algorithm before computing attention to reduce 97\% of the number of tokens in key and value matrices while maintaining a full-resolution query. We also conduct a complete theoretical analysis for low-complexity guarantees and approximation quality bounds for our SAT.
Extensive benchmarks and evaluations demonstrate that SAT outperforms all recent state-of-the-art methods,  further validating the superiority of our proposal.

\section*{Acknowledgements}

This work was supported by the National Research Foundation of Korea(NRF) grant funded by the Korea government(MSIT)(RS-2025-00573160); the “Advanced GPU Utilization Support Program” funded by the Government of the Republic of Korea (Ministry of Science and ICT); and the IITP(Institute of Information \& Coummunications Technology Planning \& Evaluation)-ITRC(Information Technology Research Center) grant funded by the Korea government(Ministry of Science and ICT)(IITP-2026-RS-2023-00259703).

The work was also supported by Hyundai Motor Chung Mong-Koo Global Scholarship to Dinh Phu Tran (1st author) and Thao Do (2nd author).

{
    \small
    \bibliographystyle{ieeenat_fullname}
    \bibliography{main}
}


\end{document}